\documentclass[USenglish]{article}

\usepackage[utf8]{inputenc}%(only for the pdftex engine)
\usepackage[small]{dgruyter}
\usepackage{dgruyter}
\usepackage{microtype}
\usepackage[style=authoryear,dashed=false]{biblatex}
\makeatletter
\AtBeginDocument{%
  \sbox\dg@wordmark{}%
}
\makeatother

\usepackage{hyperref}
\hypersetup{
    colorlinks=true,
    allcolors=black,
    urlcolor=blue,
    pdftitle={Automatic Thematic Indexing of Large Literary Corpora: A Machine Learning Approach to Voltaire's Complete Works},
    pdfpagemode=FullScreen,
    }

\begin{document}

  \articletype{Research Article}

  \author[1]{Miguel Arana-Catania}
  \author*[2]{Gillian Pink}
  \author[1]{Glenn Roe} 
  % \runningauthor{}
  \affil[1]{University of Oxford, UK}
\affil[2]{University of Oxford, UK}
  \title{\Large Automatic Thematic Indexing of Large Literary Corpora: A Machine Learning Approach to Voltaire's Complete Works}
  \runningtitle{Automatic Thematic Indexing of Large Literary Corpora}
  % \subtitle{...}

  \abstract{\small Thematic indexing --- the practice of assigning structured conceptual labels to sections of text --- is essential to scholarly access in large-scale literary and historical editions, yet it remains a largely manual, labour-intensive process. This paper explores the application of machine learning to automatic thematic indexing, using two substantial sub-corpora of the \textit{Complete Works of Voltaire} as a test case: the \textit{Essai sur les m\oe urs et l'esprit des nations} and the \textit{Questions sur l'Encyclopédie}. The task is framed as a multi-label classification problem, in which a model must assign the set of index entries that a professional indexer would apply to a given page of text. We compare four encoder-based models with classification heads and six generative large language models (LLMs) fine-tuned via Low-Rank Adaptation (LoRA). The model sizes span from approximately 3 to 120 billion parameters. Our best-performing model, from the Mistral family in a 4-bit quantised configuration, achieves F1 scores of up to 0.67; we argue that these figures represent lower bounds, although we have not measured the magnitude of the possible difference, given the inherent subjectivity of professional indexing and the frequency with which model predictions prove semantically valid despite diverging from the print index. We further evaluate cross-corpus generalisation and conduct a detailed qualitative analysis of model behaviour on literary and rhetorical features of the source texts that prove particularly resistant to automated treatment. Our findings have implications for the broader challenge of providing structured thematic access to large-scale literary and historical corpora.}
  % \classification[PACS]{...}
  % \communicated{...}
  % \dedication{...}
  % \received{...}
  % \accepted{...}
  % \journalname{Open Information Science}
  % \journalyear{2026}
  % \journalvolume{..}
  % \journalissue{..}
  \startpage{1}
  % \openaccess
  % \aop
  % \DOI{...}

\maketitle

\small

\section{Introduction}\label{introduction}

One of the founding assumptions of digital editions, and of digitised text more generally, is that they are inherently \textit{accessible} --- not only because digital channels have dramatically expanded their availability, but because they are, in a word, searchable. Yet the equation of searchability with access deserves scrutiny. Full text search, for all its power, offers a surprisingly narrow mode of engagement with a textual corpus: it presupposes that the reader already knows what terms to look for. The user arrives with a query in hand; the text responds or it does not. Discovery, in any richer sense, is left largely to chance \parencite{whitelaw2012towards}.

Traditional scholarly publishing, by contrast, has long relied on a rather different instrument. A well-constructed analytical index is not a concordance, nor a finding aid in any merely mechanical sense: it is an interpretative map of a text's contents, compiled by a professional indexer, often with specialist knowledge, who reads the work in its entirety, identifies key concepts and headwords, and progressively refines a structure of categories and sub-categories as the work takes shape. The result is qualitatively different from what can be achieved using a search function --- a structured invitation to lateral reading, browsing, conceptual navigation, and the kind of serendipitous discovery that no keyword query can reliably produce \parencite{stephen2009print, hjorland2018indexing}.

For large corpora, however, manual indexing quickly becomes either prohibitively expensive or simply impractical. The \textit{\OE uvres complètes de Voltaire / Complete Works of Voltaire} (OCV) \parencite{voltaire_ocv} presents an instructive case in point. Published by the Voltaire Foundation over fifty-five years, the edition runs to 205 print volumes and approximately fifteen million words, encompassing the full range of Voltaire's literary, philosophical, historical, and polemical output \parencite{cronk2023rediscovering}. The sheer scale of the undertaking creates a genuine problem of access: most individual works within the corpus remain unknown to non-specialists, and even lifelong readers of Voltaire can struggle to navigate the full breadth of his writings without structured aids to guide them. Now that the editorial project is complete and a state-of-the-art online edition has been launched,\footnote{\url{https://www.ov-vf.com/}} the question of how to make this body of text genuinely accessible, not merely searchable, has become newly relevant \parencite{Barker_Cronk_Roe_2026}.

Automatic indexing, broadly understood as text categorisation or classification, offers a potential response to this challenge. It is a well-established, if technically demanding, task within the broader field of supervised machine learning \parencite{sebastiani2002, dhar2021text}. In its most common formulation, a classification algorithm is trained to recognise the characteristics associated with target index terms, drawing usually on measures of lexical or semantic similarity within a manually indexed document set; it then applies what it has learned to assign terms to previously unseen texts. The capacities of such systems have been considerably extended in recent years, first by transformer-based pre-trained language models \parencite{devlin2019} and more recently by parameter-efficient methods for fine-tuning large generative models \parencite{hu2022lora, dettmers2023}. The approach nonetheless remains best suited to domains where training data is abundant and documents are relatively homogeneous in length, register, and subject matter.

The application of such methods to more complex texts, however, is less straightforward \parencite[p.~250]{duncan2022}, and literary and historical corpora give rise to particular difficulties. Collections like the OCV inhabit an unusually uneven information space: individual documents range from a handful of lines of verse to several hundreds of pages of historiographical prose, and the semantic range of the corpus as a whole --- spanning \textit{belles-lettres}, philosophical polemic, private correspondence, physics, and biblical criticism --- far exceeds that of the standard benchmarking corpora on which most text classification systems have been developed and evaluated. The nature of the target labels compounds these difficulties. The headwords of a scholarly analytical index record the interpretative judgements of an expert reader, and they are distributed with an extreme long tail in which most terms are applied only a handful of times. Existing work on automatic indexing has for the most part treated the task as one of term extraction from the text itself \parencite{csomai2008}, or has addressed controlled vocabularies designed for library cataloguing \parencite{golub2019}. Although fine-tuned language models have begun to be applied to multi-label classification \parencite{ma2025large}, they have not, to our knowledge, been tested against the professionally compiled analytical indexes of a large literary and historical corpus. It therefore remains an open question whether current models can learn to approximate this kind of interpretative labelling, under what conditions they do so most effectively, and what their failures reveal about the nature of the task.

This paper addresses that question through a series of experiments on two substantial substantial sub-corpora of the OCV: the nine-volume \textit{Essai sur les m\oe urs et l'esprit des nations} (EM) and the eight-volume \textit{Questions sur l'Encyclopédie} (QE). Both works are furnished with detailed analytical indexes in the print edition, which we use as ground truth. The task is framed as closed-vocabulary multi-label classification: given a page of text from either work (the page being retained as the unit of analysis for compatibility with the print indexes), can a model assign the set of index headwords that a professional indexer applied? Within this framing we pursue four research questions. We ask first how far supervised models can reproduce the indexer's assignments at all. We then ask how the choice between encoder-based classifiers and fine-tuned generative models, and the scale of those models, affects performance on a sparse and heavily skewed label space. We further ask whether a model trained on the index of one work can usefully be transferred to another. Finally, we ask which features of the source texts prove most resistant to automated treatment, and why.

In pursuing these questions, the paper makes four contributions. The first is a formulation of scholarly thematic indexing as a supervised classification task grounded in existing professional indexes, which allows those indexes to serve simultaneously as training signal and evaluation standard. The second is a staged comparison of encoder-based and generative architectures across a wide range of model sizes, together with an assessment of the effects of label-frequency filtering and of several variations in training procedure. The third is a cross-corpus evaluation that offers an initial measure of how well such models generalise beyond the work on which they were trained. The fourth, which we regard as the most substantive, is a qualitative analysis of model behaviour carried out by Voltaire specialists, which examines the rhetorical and literary features of the source texts that quantitative metrics alone cannot capture. The code and trained models are released openly to support further work on other corpora.

The paper proceeds as follows. Section~\ref{sec:related_work} situates the work within related research on automatic text classification, multi-label learning, and computational indexing. Section~\ref{sec:data} describes the datasets and preprocessing pipeline. Section~\ref{sec:methodology} details our methodology and presents 
experimental results across three phases of model comparison. Section~\ref{sec:discussion} discusses the implications of our findings for humanistic text collections more broadly, and Section~\ref{sec:conclusions} concludes.

\section{Related Work}\label{sec:related_work}

The task of automatically generating thematic indexes for literary and historical texts sits at the intersection of several active research areas: automatic text classification, multi-label learning, computational indexing, and the application of natural language processing (NLP) to historical text collections. We review each in turn, situating our contribution within the current state of the art.

\subsection{Automatic Text Classification}
 
Text classification --- the assignment of predefined category labels to documents --- has been a central problem in information retrieval and natural language processing for several decades \parencite{kadhim2019survey}. The field's methodological foundations are surveyed comprehensively in \parencite{sebastiani2002}, which covers document representation, dimensionality reduction, classifier construction, and evaluation, and remains an essential point of reference. The dominant paradigm through the early 2000s relied on bag-of-words representations combined with classifiers such as support vector machines (SVMs), na\"ive Bayes, and k-nearest neighbours \parencite{manning2008}. These approaches achieved strong results on standard benchmarks --- Reuters newswire, 20 Newsgroups --- but were typically designed and evaluated for single-label or binary classification settings, where each document belongs to one category or must be distinguished from one other.

The advent of deep learning brought a fundamental shift in text classification methodology. Convolutional and recurrent neural networks demonstrated the capacity to learn hierarchical feature representations directly from raw text, dispensing with hand-crafted feature engineering in favour of end-to-end training \parencite{kim2014, zhang2015}. The more consequential transformation came with the introduction of transformer-based pre-trained language models: beginning with BERT \parencite{devlin2019}, a new paradigm emerged in which large models pre-trained on massive unlabelled corpora are subsequently fine-tuned on task-specific data. This transfer learning approach has yielded state-of-the-art results across a wide range of NLP benchmarks, text classification prominently among them \parencite{sun2019, howard2018}.

\subsection{Multi-Label Classification}
 
In many real-world classification scenarios, however, including the thematic indexing task addressed in this paper, each instance may be associated with multiple labels simultaneously. Thus, multi-label classification introduces distinctive challenges beyond those of standard single-label settings: label co-occurrence dependencies, class imbalance, and the combinatorial explosion of possible label sets. \textcite{tsoumakas2007} provide an influential overview of the field, distinguishing between problem transformation methods, which reduce multi-label classification to a set of binary classification tasks, and algorithm adaptation methods, which modify existing classifiers to handle multiple labels natively. A more recent and comprehensive treatment is offered by \textcite{gibaja2015}, covering evaluation measures, label correlation modelling, and scalability considerations.

A persistent challenge in multi-label classification is the long-tailed distribution of labels: in most real-world datasets, a small number of labels account for a disproportionate share of instances while the majority remain rare \parencite{huang2021balancing}. This problem is particularly acute in humanistic indexing, where the analytical vocabulary is both extensive and finely granular. Our own datasets illustrate the point clearly: 75\% of labels in the QE index appear fewer than seven times, and 51\% of EM labels fewer than three times. Strategies for addressing label imbalance include weighted loss functions, label-aware sampling, and the selective removal of low-frequency labels --- all approaches explored in our experiments.

\subsection{Back-of-Book Indexing and Subject Indexing}

The automatic generation of indexes has been approached from two related but distinct traditions, which it is useful to separate at the outset. Subject indexing, as developed in library and information science, characterises a whole document by means of terms, usually drawn from a controlled vocabulary, so that the document can be retrieved from within a larger collection. Back-of-book indexing operates instead within a single work, producing a structured and alphabetised guide to the names, places, and concepts treated in the text and linking each to the specific locations at which it is discussed. The analytical indexes of the OCV belong to the second of these traditions, whereas the formulation of the task adopted in this paper borrows from the first. We therefore review each in turn before situating our approach in relation to both.

\subsubsection{Professional Back-of-Book Indexing}

The practice of back-of-book indexing is codified in international standards and professional manuals \parencite{browne2007indexing, weinberg2009indexing, mulvany2009indexing}, and its long history has recently been traced by \textcite{duncan2022}. These sources describe a broadly consistent procedure. The indexer reads the entire text, typically in page proofs, identifies passages that offer substantive information, and selects the headings under which a reader is likely to look for it. Those headings are then organised into main entries and subheadings, connected by cross-references, assigned locators, and finally edited for consistency and balance across the whole.

Two features of this practice are of particular relevance here. The first is that the professional indexer indexes concepts rather than simply words. A heading may legitimately be assigned to a passage in which the term itself never appears, and passing mentions that offer the reader nothing of substance are deliberately excluded \parencite{mulvany2009indexing}. The second is that the resulting index is a product of interpretative judgement. \textcite{da2021analyse} analyses the index as a paratext whose construction constitutes a genuine semiotic creation on the part of the indexer, and empirical comparison of indexes compiled by authors and by professional indexers has shown that they can differ substantially in size and structure \parencite{diodato1991back}. Accordingly, the quality of an index is assessed in this tradition against criteria that apply to the index as a whole --- completeness, navigability, and accuracy --- with no expectation of a single correct answer \parencite{bartmess2026can}.

\subsubsection{Automatic and Semi-Automatic Back-of-Book Indexing}

Research on automating this process has been comparatively sparse. Most work treats it as a problem of derived indexing and decomposes it into several stages: the extraction of candidate terms from the text, the selection of those worth indexing, the assignment of locators and, in some systems, the organisation of terms into a hierarchy. Early statistical approaches \parencite{chang1992corpus} were followed by systems grounded in computational terminology. The most developed of these is IndDoc \parencite{nazarenko2005building, zargayouna2006inddoc}, which generates a draft index from terminological analysis of the document and leaves the indexer to validate it through a dedicated interface; this semi-automatic model keeps the professional at the centre of the workflow. \textcite{csomai2006creating} established an evaluation testbed of digitised books paired with their original indexes. They subsequently showed that supervised learning with linguistically motivated features outperforms unsupervised keyphrase extraction, and that the task requires attention to both named entities and conceptual terms \parencite{csomai2008}. \textcite{wu2013can} proposed a training-free alternative explicitly modelled on the indexer's practice, in which candidate terms are ranked by an indexability score combining the structure of the book with a context-aware measure of term informativeness derived from Wikipedia. Comparable pipelines have since been developed for other languages, including Polish \parencite{marciniak2021terminology}.

Evaluation in this line of work has generally measured the overlap between automatically selected terms and the headings of an existing professional index. The modest agreement typically reported is itself informative, since it suggests how far the indexer's selection departs from what surface salience in the text would predict. The arrival of generative large language models (LLMs) has renewed interest in the question, principally among professional indexers themselves. \textcite{izzard2024generative} examined the performance of generative AI on a range of indexing tasks. Two evaluations conducted for the American Society for Indexing found that LLM-generated book indexes fell well short of professional standards of completeness, navigability, and accuracy \parencite{bartmess2025llm}. A follow-up study found that a newer generation of models under-indexed less severely, but continued to produce excessive subheadings and to fail a standard accuracy test \parencite{bartmess2026can}. It should be noted that these studies assess general-purpose models prompted to produce an entire index from scratch, a considerably broader task than the one addressed here.

\subsubsection{Automatic Subject Indexing}

The foundational framework for subject indexing is laid out by \textcite{lancaster2003}, who distinguishes between assigned indexing, in which terms are drawn from a controlled vocabulary, and derived indexing, in which terms are extracted from the text itself. \textcite{golub2019} reviews the state of automated subject indexing, examining the relationship between automatic classification and controlled vocabularies such as thesauri and classification schemes, and identifying the persistent gap between system-generated and professionally curated indexing. The evaluation of such systems has received sustained methodological attention \parencite{golub2016framework}. This attention reflects a long-standing recognition that agreement between human indexers is itself imperfect, so that a single human reference constitutes a demanding and in some respects arbitrary standard. Automatic subject indexing is now deployed operationally in national and research libraries, for example through the open-source Annif toolkit \parencite{suominen2019annif}, and the recent LLMs4Subjects shared task \parencite{d2025semeval} signals a turn towards LLMs in this domain as well.

Our approach draws on both traditions. From back-of-book indexing we take our material: professionally compiled analytical indexes, their headwords, and their page-level locators. From subject indexing we take our formulation. The full set of possible headings is defined in advance by the print index, each page is treated as a document to be characterised, and the task becomes one of assigned indexing, framed as closed-vocabulary multi-label classification. This hybrid position has consequences for the kind of comparison with professional practice that our results permit. In procedural terms, our models reproduce a single stage of the indexer's work, namely the assigning of existing level 1 headings to locations. The selection of the vocabulary itself, the construction of subheadings and cross-references, and the final editorial shaping of the index are all given in advance by the print edition. In terms of outcomes, our evaluation therefore measures agreement with a professional indexer on locator assignment. It does not address the completeness or navigability of an index as a whole, and our results should accordingly be read alongside the evaluations of whole-index generation reviewed above. The advantage of this formulation is that it allows the professional index to serve simultaneously as training signal and evaluation standard, making it possible to measure agreement with expert practice directly and at scale across several thousand pages.

\subsection{NLP for Historical Texts}
 
The application of NLP methods to historical texts has grown substantially in recent years, driven by the increasing availability of digitised collections and by advances in language model capabilities \parencite{piotrowski2012natural}. A significant strand of this work has focused on named entity recognition (NER) and linking in historical documents. The HIPE shared tasks \parencite{ehrmann2020, ehrmann2022}, organised within the CLEF evaluation forum, established multilingual benchmarks for NER on historical newspapers and foregrounded a set of challenges specific to such texts: optical character recognition (OCR) noise, archaic orthography, evolving named entities, and the scarcity of large annotated training sets. These challenges are broadly analogous to those we encounter in working with eighteenth-century French literary prose, where spelling conventions, editorial apparatus, and the density of historical and geographical reference all complicate the classification task --- though our setting adds a further layer of difficulty in the form of the long-tailed, semantically fine-grained label space described above.

Beyond NER, classification and clustering methods have been applied to a range of historical text analysis tasks. Genre classification has received sustained attention, with \textcite{underwood2019} demonstrating that machine learning models can track the long-term evolution of literary genres across large diachronic corpora. Topic modelling has similarly been applied productively to historical collections \parencite{blei2003, roe2016discourses, schoch2017}, though its sensitivity to the orthographic instability and lexical heterogeneity of pre-modern texts has prompted ongoing methodological reflection. Authorship attribution represents a further well-established classification task with a substantial literature in historical and early modern settings \parencite{koppel2009computational}, while historical text normalisation --- the standardisation of archaic and variant spellings as a preprocessing step --- has emerged as a critical enabler of downstream classification performance \parencite{bollmann2019large}. These studies collectively demonstrate that while standard NLP pipelines can be productively applied to historical texts, some form of domain adaptation --- whether through specialised training data, fine-tuning, or task-specific architectural choices --- is typically necessary to achieve acceptable performance. Our own task inherits all of these difficulties while adding one further feature: the label space we work with is not a general-purpose taxonomy but an 
analytically refined scholarly index, the categories of which reflect interpretative judgements that no off-the-shelf model has been trained to replicate.

\subsection{LLM Fine-Tuning for Classification}
 
The emergence of large language models with tens or hundreds of billions of parameters has opened new avenues for text classification, particularly through instruction tuning and parameter-efficient fine-tuning methods. The LoRA framework \parencite{hu2022lora} introduced the concept of injecting trainable low-rank decomposition matrices into the layers of a frozen pre-trained model, dramatically reducing both the number of trainable parameters and the memory footprint of fine-tuning without sacrificing downstream task performance. QLoRA \parencite{dettmers2023} extended this approach by combining LoRA with 4-bit quantisation, making it possible to fine-tune models with up to 65 billion parameters on consumer-grade hardware --- a development that has substantially lowered the barrier to entry for research groups without access to large-scale graphics processing unit (GPU) infrastructure.

These techniques have been widely adopted for adapting LLMs to domain-specific tasks, including multi-label classification \parencite{ma2025large}. In our experiments, we employ LoRA-based fine-tuning of quantised generative models, treating the models as text decoders that produce the predicted label set directly as a structured text string --- a generative formulation that contrasts with the more conventional encoder-plus-classifier-head architecture. Our comparative results suggest that, for the particular challenges of thematic indexing --- large label spaces, low-frequency labels, semantically rich and generically diverse texts --- this approach offers meaningful advantages, a finding we return to in the discussion.

The present study represents, to our knowledge, one of the first systematic attempts to apply LLM fine-tuning to the automatic generation of analytical indexes for a large-scale literary and historical corpus. While the individual components of our approach --- multi-label classification, LoRA-based fine-tuning, historical text processing --- are each well established in their respective literatures, their combination in the service of scholarly indexing has not previously been attempted. The results offer both practical guidance for digital edition projects confronting the accessibility problem outlined above, and methodological contributions to the broader field of computational text analysis in the humanities.

\section{Dataset and Preprocessing}\label{sec:data}

The two texts selected for this study are Voltaire's \textit{Essai sur les m\oe urs et l'esprit des nations} (EM) and \textit{Questions sur l'Encyclopédie} (QE). Both are works of non-fiction: the former is a universal history of customs and nations, the latter a collection of alphabetically ordered articles in the encyclopaedic tradition. As discussed in Section \ref{sec:discussion}, however, both make extensive use of rhetorical and fictional devices, and both include detailed analytical indexes produced as part of the print edition. For EM, seven volumes were used --- numbered 22, 23, 24, 25, 26A, 26B, and 26C in the OCV sequence --- omitting volumes~21 and 27, which contain a scholarly introduction and appendices respectively. For QE, seven volumes 
were likewise used (38--43, excluding volume 37, which contains the scholarly introduction).

Both texts were available in TEI-XML format; preprocessing involved extracting the main text and stripping all markup tags. The analytical indexes were available as structured files recording, for each entry, its hierarchical level and the list of volumes and pages to which it applies, including footnotes. QE contains four hierarchical levels, with 419 \textit{index1}, 1,237 \textit{index2}, 216 \textit{index3}, and 4 \textit{index4} entries; EM contains three levels, with 556 \textit{index1}, 1,972 \textit{index2}, and 230 \textit{index3} entries. Individual entries may in turn reference multiple page ranges (e.g. \ \textit{``Acre, iii.334, 355, 356, 366, 368, 377, 382n, 390; conquise par Saladin, iii.326--29; prise d', iii.335--36, 337n''}), so the total number of indexable label instances exceeds the number of entries. This study focuses throughout on level~1 entries unless otherwise specified. Cross-references (entries preceded by \textit{voir}) and entries referring exclusively to footnotes were excluded, the latter because our analysis operates at the level of the page rather than the note. The indexes were converted from their standard format (e.g.\ \textit{``abeilles, ii.39--46, v.370, 380, 450, vii.449, 475''}) into a label format suitable for machine learning training (e.g.\ \textit{``abeilles,2,39; abeilles,2,40;..."}), producing datasets in which each page of text is associated with the set of index labels assigned to it by the professional indexer --- a multi-label classification task in any standard sense.

The datasets were divided into training, validation, and test subsets in an approximate 80/10/10 ratio, using an iterative stratification method 
designed for multi-label data \parencite{szymanski2017network, sechidis2011stratification} as implemented in the Scikit-multilearn library \parencite{szymanski2019scikit}, in order to achieve the best possible distributional balance across subsets. For QE, this yielded 2,279/365/368 training/validation/test samples across 419 labels; for EM, 1,684/258/241 samples across 866 labels. The validation sets were used for hyperparameter tuning of the models. All the numerical results reported are from the test sets. None of the methodological decisions made in this project (hyperparameter values, architectural choices, model selection) was based on the results obtained using the test sets.

\begin{figure}[htp!]
    \centering
    \includegraphics[width=0.49\linewidth]{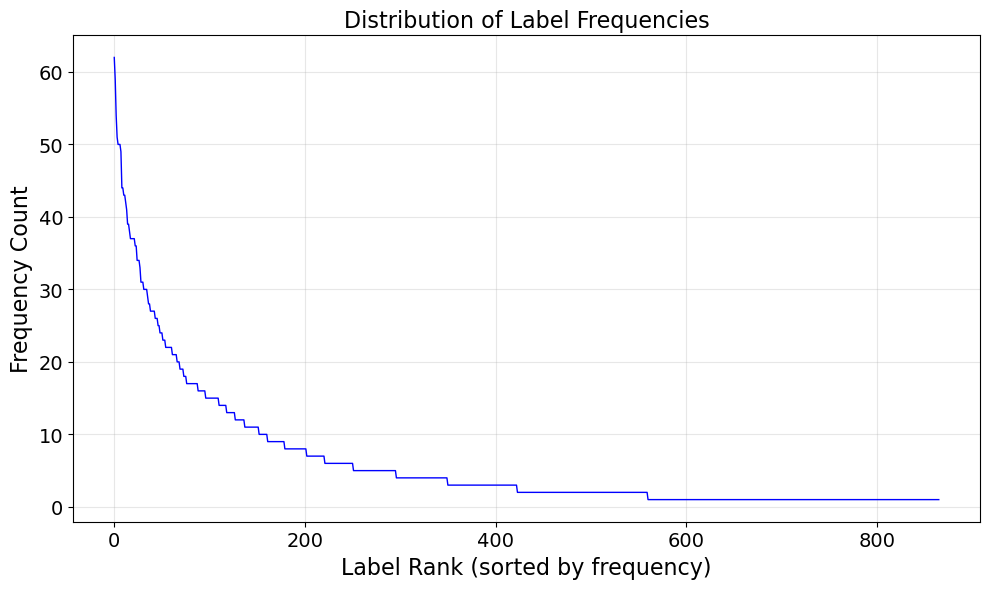}
    \hfill
    \includegraphics[width=0.49\linewidth]{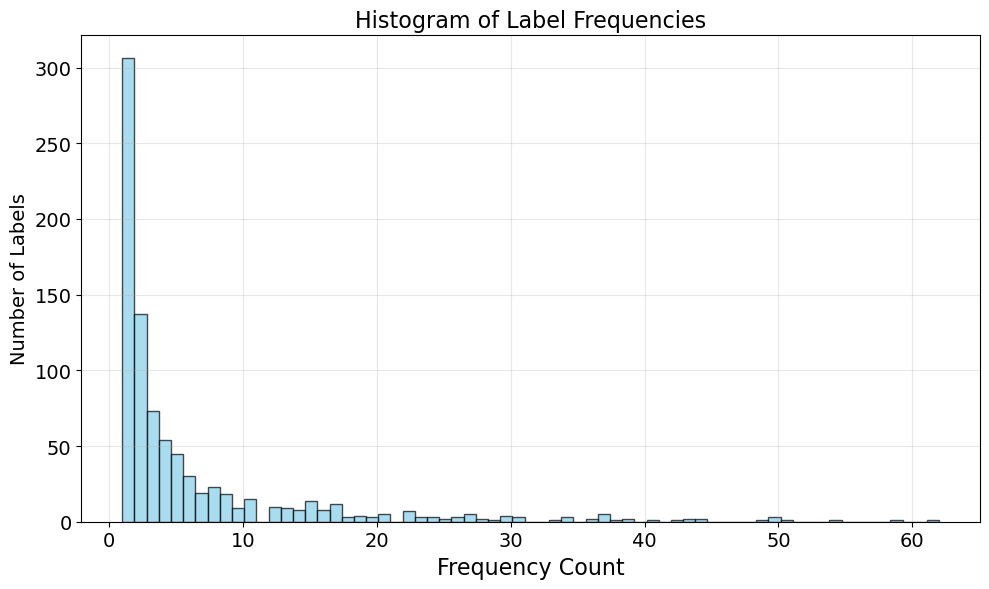}
    \caption{Label frequencies for the EM dataset.}
    \label{fig:distr_labels_EM}
\end{figure}

\begin{figure}[htp!]
    \centering
    \includegraphics[width=0.49\linewidth]{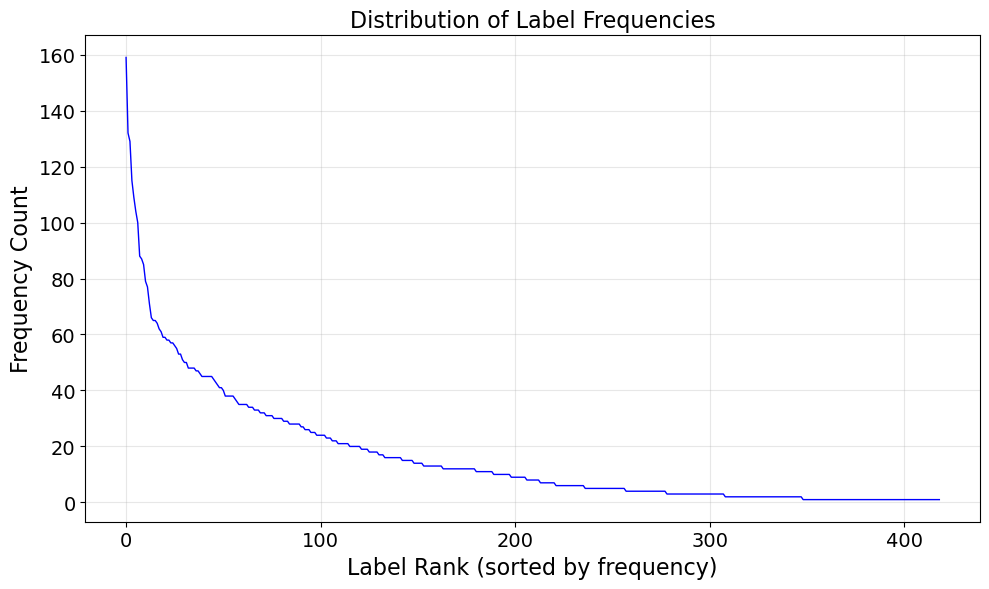}
        \hfill
    \includegraphics[width=0.49\linewidth]{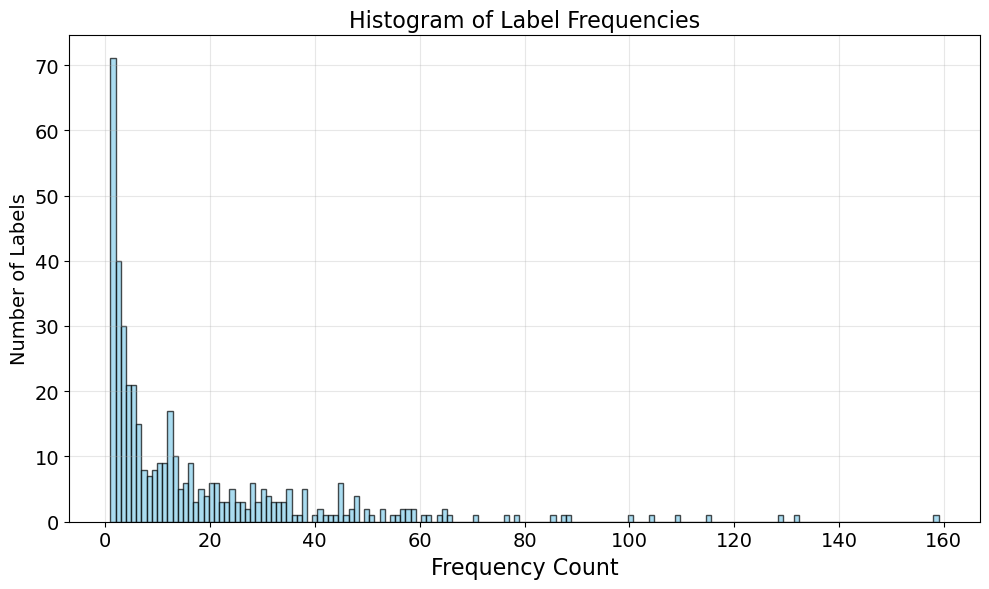}
    \caption{Label frequencies for QE the dataset.}
    \label{fig:distr_labels_QE}
\end{figure}

A defining characteristic of both datasets --- and a central challenge for model training --- is the extreme skew in label frequency. In QE, 75\% of labels appear fewer than 7 times across the entire corpus; in EM, 75\% appear fewer than 23 times. At the page level, the distributions are similarly sparse: in both datasets, approximately 37\% of pages carry a single label, and around 80\% carry 3 or fewer. Figures~\ref{fig:distr_labels_EM} and~\ref{fig:distr_labels_QE} illustrate these distributions. This degree of label sparsity poses a substantial challenge for supervised learning: for the majority of index terms, the model has access to only a handful of positive training examples from which to learn reliable classification criteria.

\section{Methodology and Results}\label{sec:methodology}

The experimental methodology is organised into three successive phases, each designed to narrow the field of candidate approaches before committing to full-scale training. In the first two phases, we compare models and architectures of increasing size to identify the most promising configuration; in the third, we conduct final training runs with the selected model and evaluate its capacity to generalise across corpora. Throughout, we report the results of additional experiments involving modifications to training procedure, dataset composition, and model configuration, in order to characterise the sensitivity of results to these choices.

The software developed during this project, as well as the machine learning models we have trained, are publicly available as open-source in the project repository\footnote{\url{https://github.com/Digital-Scholarship-Oxford/digitally-indexing-voltaire}}.

\subsection{Comparison of Medium-Sized Models}\label{sec:med_ml}

The first phase of comparison was conducted on the EM dataset using models of approximately 5 billion parameters. Two broad architectural approaches were explored in sequence: encoder-based models with classification heads, and generative decoder models producing labels directly as text output. These two approaches are the most common architectures for classification problems in NLP.

For the encoder-based approach, pre-trained LLMs were used as frozen feature extractors, with a series of fully connected linear layers added on top. The final layer contained one neuron per label, yielding independent probability estimates for each class. Four models were evaluated: Qwen3-4B, Llama-3.2-3B, Gemma-3-4B, and Mistral-7B-v0.3, using the following Hugging Face checkpoints respectively: \texttt{Qwen/Qwen3-Embedding-4B}, \texttt{meta-llama/Llama-3.2-3B}, \texttt{google/gemma-3-4b-pt}, and \texttt{mistralai/Mistral-7B-v0.3}. These models belong to the state-of-the-art families of language models. The only constraint we have taken into account when selecting the models is the ability to carry out training and inference on a workstation with a single GPU, as these were the resources available for this project. The classification head consisted of stacked Linear--ReLU--Dropout blocks, with the first layer sized proportionally to the encoder output by a scale factor hyperparameter, and subsequent layers reducing in size by a factor of $2^{(i+1)}$, where $i$ denotes the layer index. Encoder weights were frozen throughout; only the classification head was trained. The loss function was weighted binary cross-entropy, implemented using BCEWithLogitsLoss, with per-class weights computed as the ratio of negative to positive counts, to compensate for label imbalance. For training, we used an AdamW optimiser with a LinearLR scheduler decaying the learning rate by a factor of 0.1 over the course of training. All models were implemented in PyTorch, using the Sentence Transformers\footnote{\url{https://sbert.net}} and Transformers\footnote{\url{https://huggingface.co/docs/transformers}} libraries. Hyperparameter tuning was conducted with Optuna,\footnote{\url{https://optuna.org}} searching over dropout rate, learning rate, weight decay, batch size, first-layer scale factor, and number of layers; models were trained for five epochs during tuning and up to 15 epochs thereafter.

To evaluate the results, we have used the following metrics: precision, recall and micro-F1. From now on, we will refer to the latter as F1. All three metrics have been implemented using the scikit-learn library\footnote{\url{https://scikit-learn.org/stable/api/sklearn.metrics.html}}. The results of this first comparison are shown in Table~\ref{tab:medium_size_ml_models_em_dataset}. Performance across the encoder-based models was modest, with F1 scores ranging from 0.24 to 0.34, reflecting the difficulty of the task and the limitations of a frozen encoder for a label space of this granularity and sparsity.

These results motivated a shift to a generative decoding formulation, in which models produce the predicted label set directly as a structured text string rather than via a classification head. This approach accommodates larger models more naturally and --- as our subsequent results confirm --- yields substantially better performance. In this second approach, we compared an OpenAI proprietary model with several open-source alternatives. The proprietary model used was GPT-4.1-mini (\texttt{gpt-4.1-mini-2025-04-14}). Given the cost of fine-tuning on texts of this length, experiments were limited to the mini variant. OpenAI does not publish parameter counts, architectural details, or training data for its models. Fine-tuning used the OpenAI JSONL format. The prompt provided in the user content field was as follows: \textit{``You are an expert in French historical text classification. You need to classify the following French text about Voltaire's historical works by assigning relevant thematic labels. Text to classify: \{text\}. Available labels: \{labels\}. Instructions: 1. Read the text carefully. 2. Identify all relevant themes present in the text. 3. Return only a Python list of label names that apply to this text. 4. Be precise --- only include labels that are clearly relevant to the content.''} Structured JSON output was enforced via a response schema specifying a \texttt{predicted\_labels} array. The model was trained for 3 epochs with automatic batch size and learning rate multiplier. The results for GPT-4.1-mini are included in Table~\ref{tab:medium_size_ml_models_em_dataset}. With an F1 of 0.44, the generative formulation outperforms all encoder-based models, confirming the value of the architectural shift.

Several modifications to the GPT-4.1-mini fine-tuning procedure were explored, none of which produced meaningful improvements. Expanding the context window to include the preceding and following pages alongside the target page yielded an F1 of 0.44. Testing different numbers of training epochs from one to five produced a marginal improvement at five epochs (F1 = 0.46). A label partitioning experiment, in which the full label set was divided into four subsets of approximately 216 labels each and four independent models trained accordingly, increased recall to 0.55 but caused precision and F1 to fall to 0.21 and 0.30 respectively --- a result consistent with the hypothesis that smaller label sets reduce precision by increasing false positive rates across the combined output. These negative results are informative in themselves: they suggest that 
the limiting factor at this scale is not context size, training duration, or label set granularity, but model capacity.

\begin{table}[htp!]
        \caption{Evaluation results of training of medium-sized machine learning models on the EM dataset.}
    \centering
        \resizebox{0.6\columnwidth}{!}{
    % \begin{tabular}{|c|c|c|c|}
    \begin{tabular}{llll}
        % \hline
         \textbf{Model} &  \textbf{Precision} &  \textbf{Recall} & \textbf{F1} \\
         % \hline
        \midrule
         Qwen3-4B & 0.4450 & 0.2724 & 0.3380 \\
         \hline
         Llama-3.2-3B & 0.2526 & 0.2308 & 0.2412 \\
         \hline
         Gemma-3-4b & 0.3624 & 0.2532 & 0.2981 \\
         \hline
         Mistral-7B-v0.3 & 0.3301 & 0.2179 & 0.2625 \\
         \hline
         GPT-4.1-mini & 0.3929 & 0.5000 & 0.4401 \\
         % \hline
    \end{tabular}
    }
    \label{tab:medium_size_ml_models_em_dataset}
% \vspace{-30pt}
\end{table}

\subsection{Comparison of Large Quantised Models}\label{sec:large_ml}

The second phase scaled up to models of approximately 15 billion parameters, evaluated on the QE dataset. To manage computational cost, all open-source models were used in 4-bit quantised form. GPT-4.1-mini was retained for cross-phase comparison. All models were used as generative text decoders with the same prompt as in Section~\ref{sec:med_ml}.

The models compared were: GPT-4.1-mini, Qwen3-14B, Gemma-3-12B, Phi-4-14B, Mistral-Small-3.2-24B, and Mistral-Large-123B, using the following Hugging Face checkpoints: \texttt{unsloth/Qwen3-14B-unsloth-bnb-4bit}, \texttt{unsloth/gemma-3-12b-it-unsloth-bnb-4bit}, \texttt{unsloth/phi-4-unsloth-bnb-4bit}, \texttt{unsloth/Mistral-Small\--3.2-24B-Instruct-2506-unsloth\--bnb-4bit}, and \texttt{unsloth/Mistral-Large-Instruct-2407-bnb-4bit}. Open-source models were fine-tuned using Low-Rank Adaptation (LoRA) \parencite{hu2022lora} with rank $r = 16$ and $\alpha = 16$, without dropout, for three epochs. Training used the TRL library's\footnote{\url{https://huggingface.co/docs/trl}} SFT Trainer with Unsloth framework integration,\footnote{\url{https://huggingface.co/docs/trl/unsloth_integration}} a learning rate of $2 \times 10^{-4}$, weight decay of 0.001, a linear scheduler, and AdamW8bit as optimiser.

It is important to emphasise that the results obtained in this case are not directly comparable with those obtained in the previous section. A number of factors have changed between the two sections: the corpus, the model architecture (including the switch from encoder-based classifiers to generative decoder models), the scale of the models, the training procedures, etc. Therefore, the conclusions drawn in each section must be understood within their respective contexts. The improved numerical results obtained here should be interpreted as indicative only. In the next section, we will carry out a comparison of results across corpora, which will enable us to interpret the effect of the change in corpus on the results.

Results are shown in Table~\ref{tab:large_size_ml_models_qe_dataset}. The Mistral family achieved the strongest performance, with Mistral-Large-123B reaching an F1 of 0.65 and Mistral-Small-3.2-24B following closely at 0.64. Given the negligible performance gap and the substantial difference in computational requirements, Mistral-Small-3.2-24B was selected for the final phase.

\begin{table}[htp!]
    \caption{Evaluation results of training of large-sized machine learning models on the QE dataset.}
    \centering
            \resizebox{0.6\columnwidth}{!}{
    % \begin{tabular}{|c|c|c|c|}
    \begin{tabular}{llll}
         % \hline
         \textbf{Model} &  \textbf{Precision} &  \textbf{Recall} & \textbf{F1} \\
         % \hline
         \midrule
         GPT-4.1-mini & 0.5763 & 0.5305 & 0.5525 \\
         \hline
         Qwen3 14B & 0.4781 &  0.6248 & 0.5417 \\
         \hline
         Gemma 3 12B & 0.3725 & 0.5213 & 0.4345 \\
         \hline
         Phi-4 14B & 0.5221 & 0.7431 & 0.6133 \\
         \hline
         Mistral Small-3.2 24B & 0.5595 & 0.7560 & 0.6431 \\
         \hline
         Mistral Large 123B & 0.5679 & 0.7652 & 0.6520 \\
         % \hline
    \end{tabular}
    }
    \label{tab:large_size_ml_models_qe_dataset}
% \vspace{-30pt}
\end{table}

As in the previous phase, a series of methodological variants were explored using the QE dataset. Combining GPT-4.1-mini and Mistral-Small-3.2-24B sequentially --- using the former's output labels as additional input to the latter for confirmation or rejection --- produced no meaningful improvement (F1: 0.5525 to 0.5558), suggesting that the sequential architecture adds latency without complementary benefit. Extending Mistral-Small-3.2-24B training to fifteen epochs produced only marginal gains (F1: 0.6431 to 0.6488), indicating that the model converges rapidly. Converting all text to lowercase prior to training, motivated by the hypothesis that capitalised proper nouns were exerting undue influence on label assignment, likewise produced no significant change (F1 = 0.6479). Finally, augmenting the dataset with \textit{index2} and \textit{index3} entries --- renamed by combining each term with its parent \textit{index1} headword to avoid expanding the label space --- produced virtually no improvement (F1: 0.6431 to 0.6457). Taken together, these results suggest that performance at this scale is relatively robust to the modifications explored, and that further gains are more likely to come from architectural choices or label space restructuring than from training procedure adjustments.

\subsection{Final Training and Out-of-Distribution Evaluation}\label{sec:final_ml}

The final phase used Mistral-Small-3.2-24B in its 4-bit quantised configuration for two experiments: an assessment of the effect of label frequency filtering, and a cross-corpus out-of-distribution evaluation.

For the label filtering experiment, low-frequency labels were removed from both datasets prior to training: in QE, labels appearing fewer than 5 times were excluded, reducing the label set from 419 to 257 (a reduction of 39\%); in EM, labels appearing fewer than three times were excluded, reducing the label set from 866 to 423 (51\%). Despite removing a large proportion of labels, the total number of indexed page instances was affected only modestly --- by 5\% for QE and 10\% for EM --- since low-frequency labels account for a small share of overall label assignments. Filtering produced consistent improvements: F1 rose from 0.64 to 0.67 for QE, and from 0.42 to 0.47 for EM. While the gains are not dramatic, they confirm that reducing the number of labels improves performance. This may be influenced by two factors: firstly, there are fewer categories to consider, so the task at hand is simpler; and secondly, labels with very few training examples, which are thus more difficult to learn during the training, are eliminated. When it comes to scholarly indices, rare concepts can be highly valuable, so if this methodology is to be applied, it is advisable to carry out a manual appraisal of uncommon tags to decide whether to exclude them.

For the out-of-distribution experiment, models trained on one corpus were evaluated on the other: the QE-trained model was applied to EM, and the EM-trained model to QE. When discussing cross-corpus inference, it is important to distinguish between the different architectures used in this project. When using models with classification architectures, such as those used in Section \ref{sec:med_ml}, this type of cross-corpus inference cannot be carried out directly, as the model can only produce the outputs defined and used during its training. In such cases, it is necessary to modify the model’s final layers to adapt them to the new set of labels and to retrain the model, at least partially. However, when using generative models such as those employed in this section, it is possible to reuse the models directly on other corpora without retraining. The possible labels for each use are defined in the model’s prompt, so it is possible to modify these between uses of the same model; furthermore, as these models produce their outputs as a text string, they are not limited to certain types of output. Of course, a model trained for one type of label will be much more efficient when used on that set (in-distribution) compared to a new set of labels (out-of-distribution), as we shall see below. But in any case, it is possible to use them directly on new sets, which can be useful in cases where the resources to retrain the models are not available.

The results, shown in Table~\ref{tab:final_training}, reveal a consistent performance drop of approximately 10--20 percentage points in F1 when moving from in-distribution to out-of-distribution evaluation. As mentioned above, this is to be expected given that the two corpora have entirely distinct label sets. Nevertheless, the results show that it is possible to reuse a model to index new corpora if we accept the low level of quality measured. This can be useful in practice if we do not have the computational resources to retrain models or the human resources to carry out manual indexing. We can thus provide researchers with an initial point of entry to a collection who would otherwise have no other way of navigating it thematically. In any case, such use must always be clearly labelled as partial automated indexing, to distinguish it from manual results produced by subject matter experts.

\begin{table}[htp!]
    \caption{In- and Out-of-distribution evaluation results for Mistral-Small-3.2-24B, 4-bit quantised version.}
    \centering
    \resizebox{0.9\columnwidth}{!}{
    \begin{tabular}{lllll}
        % \hline 
        \textbf{Training Details} & \textbf{In/Out} & \textbf{Precision} &  \textbf{Recall} & \textbf{F1} \\
        % \hline
        \midrule
        Trained on EM, evaluated on EM & In & 0.3339 & 0.6186 & 0.4337 \\
        \hline
        Trained on QE, evaluated on EM & Out & 0.2554 & 0.4135 & 0.3158 \\
        \hline
        Trained on QE, evaluated on QE & In & 0.5595 & 0.7560 & 0.6431 \\
        \hline
        Trained on EM, evaluated on QE & Out & 0.3801 & 0.5656 & 0.4547 \\
        % \hline
    \end{tabular}
    }
    \label{tab:final_training}
% \vspace{-30pt}
\end{table}

\section{Discussion}\label{sec:discussion}

The quantitative results presented above warrant closer examination before their implications are drawn out. In so doing, we follow a well-established principle in the evaluation of automatic indexing, according to which comparison with a single gold standard is insufficient on its own. \textcite{golub2016framework} distinguish between the direct evaluation of indexing quality, either through comparison with a gold standard or through assessment by expert evaluators, the evaluation of computer-assisted indexing within an operational indexing workflow, and the indirect evaluation of indexing through retrieval performance. They argue that the difficulties inherent in judgements of aboutness and relevance make it necessary to combine more than one of these approaches, a conclusion reinforced by the long-standing observation that agreement between human indexers is itself imperfect \parencite{funk1983indexing}. The quantitative results reported in Section \ref{sec:methodology} correspond to the first of these approaches, measured against the print index as the gold standard. To complement them, we carried out a direct evaluation by expert assessment: a researcher specialised in Voltaire reviewed a random sample of the model outputs, and the resulting assessment was in turn reviewed by a second expert in the field. This assessment combines an analysis of success, identifying predictions that are valid despite diverging from the print index, with an analysis of failure, identifying the textual features that lead the model astray. Following \textcite{golub2016framework}, we regard both as essential to understanding what the quantitative metrics do and do not capture. Workflow-based and retrieval-based evaluation, which would assess the usefulness of the model's output to indexers and to readers respectively, lie beyond the scope of the present study, and we return to them among the directions for future work.

A further observation that is raised is that the evaluation framework systematically understates model performance in ways that are worth making explicit. A predicted label was counted as false whenever it diverged from the human-attributed label derived from the print index --- regardless of its semantic appropriateness. This means that a model prediction can be both plausible and penalised simultaneously. One such example is a page from the QE article \textit{``Abeilles''} in which Voltaire considers the social and economic organisation of a bee colony. The human indexer assigned the labels \textit{``abeilles''} and \textit{``anthropomorphisme''}; the model adds \textit{``science expérimentale''}, which is entirely pertinent, since Voltaire discusses the close observations of his contemporary Jean-Baptiste Simon over a period of twenty years \parencite[p.~41]{voltaire_ocv_38}. Similarly, a page of the article \textit{``Superstition''} defends Anthony Ashley Cooper, third Earl of Shaftesbury, and his grandfather the first Earl, from recent literary attacks, contrasting their quiet lifestyle with the assassination of kings carried out or attempted by religious fanatics. The human indexer attributed four labels to this page \parencite[p.~319]{voltaire_ocv_43}: \textit{``philosophie''}, \textit{``raison humaine''}, \textit{``superstition''}, and \textit{``régicide''}; the model produces the first two, but substitutes \textit{``société civile''} for \textit{``régicide''}. It could reasonably be argued that regicide is a subcategory of civil society --- and while the substitution loses a degree of semantic specificity, \textit{``société civile''} is far from wrong. It must always be borne in mind that thematic indexing, however expertly conducted, is inherently subjective: no two professional indexers will produce precisely the same labels or structure for a text of any length or complexity.

In places, the model produces results that compare favourably with the print index. In chapter~81 of EM, which addresses customs and lifestyle in thirteenth- and fourteenth-century Europe, the human indexer attributes only the term \textit{``courage''} to a short page that nevertheless contains a thematically rich conclusion, comparing the relative poverty of France in the period with the riches of the Italian mercantile city-states, where convenience and opulence reigned. These 
advantages, Voltaire adds, gave rise to genius --- as they do to courage \parencite[p.~267]{voltaire_ocv_24}. The model has identified \textit{``courage''} but more pertinently also includes \textit{``Italie''} and \textit{``luxe''} (luxury). That France, mentioned more peripherally than Italy, does not feature as a label is particularly encouraging.

\begin{figure}[htp!]
    \centering
    \includegraphics[width=0.45\columnwidth]{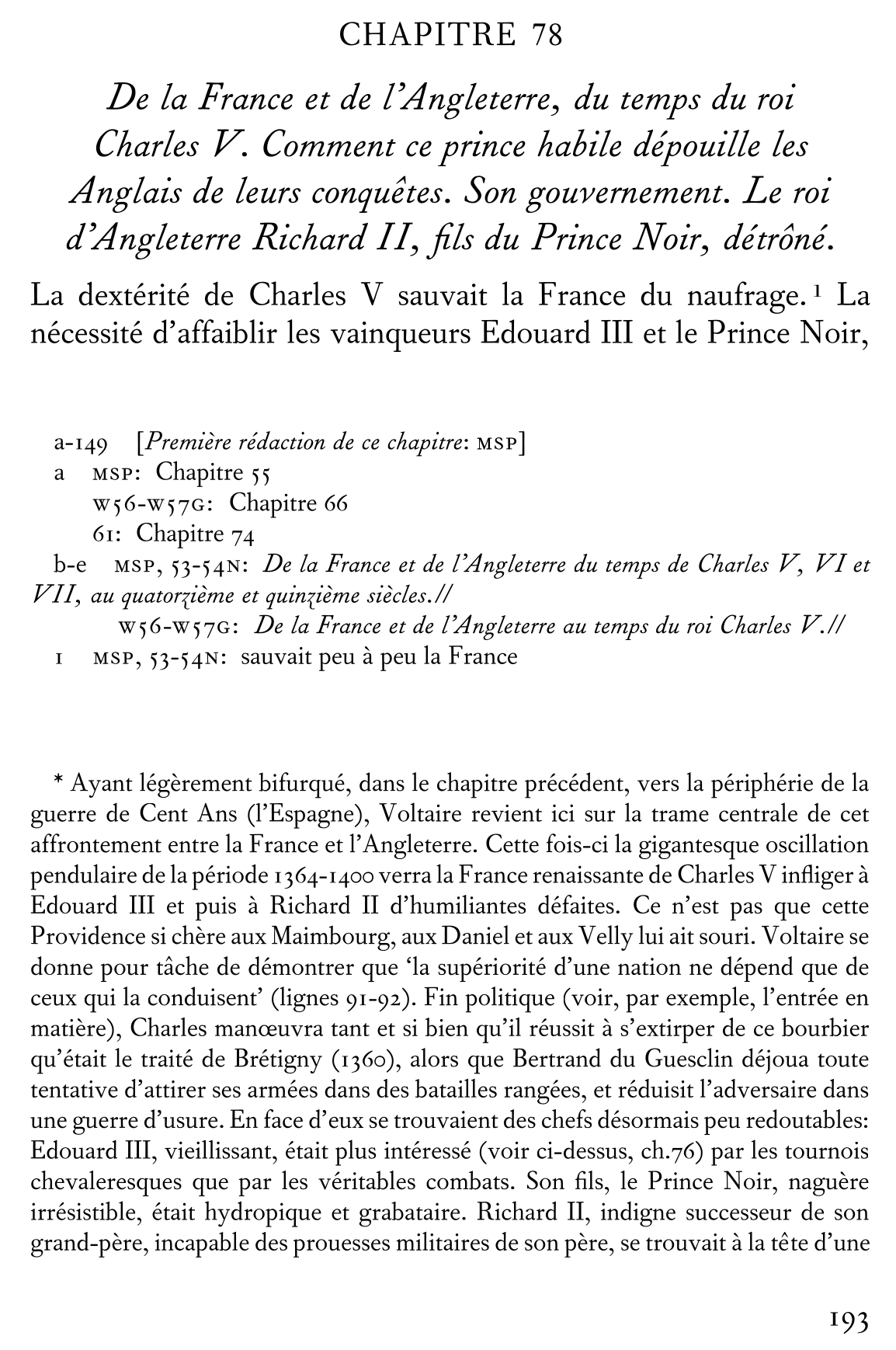}
    \caption{Example of a page of EM containing a small amount of primary text.}
    \label{fig:ocv_page}
\end{figure}

The effect of page length on model performance is worth examining. The page just discussed comes at the end of a print chapter, with blank space beneath the final lines of text; it contains only fifty-nine words across 7 short lines. Other pages carry relatively little primary text owing to the space taken up by lengthy editorial notes and variant readings (see Figure~\ref{fig:ocv_page}). A shorter text might be expected to produce less reliable labelling. When results are sorted by word count, however, 
this hypothesis is not consistently borne out. Among the twenty shortest EM pages in the results, four receive labels that match those of the human indexer exactly, and a further nine at least partially overlap or constitute plausible alternatives. In a few cases the presence of a chapter title at the head of the page --- which would have been ignored by the indexer --- accounts for certain model-predicted labels: the opening of EM chapter~5, entitled 
\textit{``De la Perse, au temps de Mahomet le prophète, et de l'ancienne religion de Zoroastre''}, predictably generates the label \textit{``zoroastrisme''} even where the body text on that page contains no reference to Zoroastrianism \parencite[p.~97]{voltaire_ocv_22}. There are also instances where the brevity of a page genuinely defeats the model: at the end of EM chapter~103, a page of twenty-four words --- \textit{``tout aux autres; les servant par avarice, les détestant par fanatisme, se faisant de l'usure un devoir sacré. Et ce sont nos pères!''} \parencite[p.~185]{voltaire_ocv_25} --- forms the tail end of a longer development on the Jewish nation, which the human indexer has placed under \textit{``Juifs''}. Without the preceding context, the model not unreasonably assigns \textit{``avarice''} and \textit{``sentiments haine''}.

Alongside results that compare positively with the print index, or are at least equal in validity, while being different, there are inevitably several that are less satisfactory. In the QE article \textit{``Supplices''} (Torture), Voltaire illustrates his argument with examples of historical figures subjected to torture following dubious accusations of poisoning \parencite[p.~332]{voltaire_ocv_43}. The print index gives the label \textit{``empoisonnement''} (poisoning), which the model also supplies, while adding \textit{``gloire''} (glory) and \textit{``imposture''} (deception). These last two terms clearly have their origins in single words that are peripheral to the main subject: at one point Voltaire addresses those who supported the accusations of poisoning but named different accomplices: \textit{``Accordez-vous donc, pauvres imposteurs''} (Get your stories straight, miserable deceivers), he exclaims. In the next paragraph, Voltaire refers in passing to the accusers being jealous of the glory of Holy Roman Emperor Charles~V. Neither word provides justifiable grounds for the labels assigned.

A particularly clear instance of the model being misled by surface features occurs in the QE article \textit{``Alexandre''}, on Alexander the Great, where Voltaire adds a dig at Alexander's questions to the Gymnosophists in India, which he describes as worthy of the \textit{Mercure galant}, the late seventeenth- and early eighteenth-century literary gazette that eventually became the \textit{Mercure de France}. The page as a whole \parencite[p.~186]{voltaire_ocv_38} concerns ancient history and 
the reliability of Plutarch's testimony, yet the model assigns it the label \textit{``galanterie''} --- evidently on the basis of the periodical title. To test whether capitalised proper nouns and titles were exerting undue influence on label assignment more generally, the full dataset was converted to lowercase and the experiment rerun, as noted in Section~\ref{sec:large_ml}, but this produced no significant change in performance.

These two cases --- one involving direct address to an imagined interlocutor, the other a humorous historical comparison --- begin to take us beyond the domain of straightforward thematic content into the territory of Voltaire's rhetoric. Voltaire is a literary writer, master stylist, and polemicist; rhetorical effects permeate his prose even when he is writing history or essays, where his voice and perspective were key selling points for contemporary readers \parencite[p.~62]{menant2021}. Although neither the EM nor the QE is a work of fiction, Voltaire's recourse to dialogue, invented interlocutors, and ironic framing brings our task close to the familiar difficulties of fiction indexing, where aboutness must be inferred from, rather than read off, the text and where agreement between indexers is correspondingly low \parencite{beghtol1990access, saarti2002consistency}. This is particularly true of QE, a compendium of articles written late in life on a wide variety of topics, many of them serious but some mischievous or playful. Like its better-known precursor the \textit{Dictionnaire philosophique} (DP), some articles are written in dialogue form. In the case of \textit{``Liberté''} and \textit{``Liberté de penser''}, the themes emerge clearly enough despite the dialogic format \parencite[p.~47; 53]{voltaire_ocv_42B}. Irony is a more intractable challenge, as are allegory, periphrasis, and comparison, all of which can lead the model away from the underlying topic. The QE article \textit{``Gargantua''} \parencite[pp.~1--5]{voltaire_ocv_42A} would have offered a particularly instructive test case: a masterclass in oblique polemic, it mounts an indirect attack on the Old Testament and biblical apologetics under the guise of a ludicrous discussion of the existence of the giant hero of Rabelais's novel (for an analysis of this article, see \parencite[pp.~351--56]{sareil1983}; see also Pink's introductory note in \parencite[p.~1]{voltaire_ocv_42A}). Unfortunately it does not feature in our test split. The DP article \textit{``Guerre''} does, however, and it offers a comparable example. The article pillories priests and the Church for celebrating bloody military victories while ordinarily preaching against vice --- a line of attack that, in the eighteenth century, required careful circumlocution. The word \textit{``prêtre''} (priest) does not appear, nor is the Bible named; Voltaire identifies his subjects through periphrasis and comic detail. His priests are referred to as \textit{``harangeurs''}, their vestments and speech described as if observed by a visitor from an alien culture. The model, thoroughly bamboozled by Voltaire's verbal acrobatics, produces the labels \textit{``démonisme''} and \textit{``nourriture''} (food), having fixed on two items in an accumulation of things the priest condemns in peacetime: \textit{``que Polyeucte et Athalie sont les ouvrages du démon; qu'un homme qui fait servir sur sa table pour deux cents écus de marée un jour de carême, fait immanquablement son salut''} (that \textit{Polyeucte} and \textit{Athalie} [plays by Corneille and Racine] are the work of the devil; that a man guarantees his salvation by having 200 crowns worth of seafood [instead of meat, is implied] served at his table on a day in Lent) \parencite[p.~191]{voltaire_ocv_36}. The labels \textit{``guerre''}, \textit{``Église catholique''}, and \textit{``religion''} were available 
in the label set but were not applied. It is worth noting that \textit{``prêtres''} was not available, since in the human-generated  index it appears as a sub-entry under \textit{``religion''} --- a reminder that the constraint of working with level~1 labels limits the model's expressive range in ways that are not always visible in the quantitative results.

The treatment of quotations presents a related but distinct difficulty. Some QE articles contain long extracts from verse. In some cases the model copes reasonably well: a section of the article \textit{``Athéisme''} reproduces almost in full Voltaire's 1769 poem \textit{Épître à l'auteur du livre des Trois Imposteurs}, a polemical work that tackles its subject directly and without heavy recourse to figurative language; the model's proposed labels closely match those of the human indexer \parencite[p.~174]{voltaire_ocv_39}. The poetry present in the text is not always Voltaire's own, however. He peppers his articles with prose and verse quotations from writers ranging from classical antiquity to his own time. The QE article \textit{``Langues''} contains a passage from Molière's \textit{L'Amour médecin}, included to illustrate the historical 
usage of the French word \textit{``rate''} rather than the more recent English borrowing \textit{``spleen''} \parencite[pp.~7--8]{voltaire_ocv_42B}; the human indexer classifies page~8 as comparative linguistics, while the model opts for \textit{``corps humain''} and \textit{``rire''} (human body and laughter). Similarly, quotations from Virgil and Seneca --- for which Voltaire helpfully provides French translations --- skew the interpretation of an article on the development of the notion of hell 
(\textit{``Enfer''}), where the presence of the word \textit{``imposture''} in Voltaire's translation generates a corresponding but non-pertinent label from the model \parencite[p.~109]{voltaire_ocv_41}. A related problem emerges in the article \textit{``Bouffon, burlesque''}, where Voltaire quotes passages from Scarron's plays \textit{Don Japhet d'Arménie} and \textit{Le Virgile travesti}: what the human indexer places under \textit{``goût''} (taste) and \textit{``style''} is assigned the label \textit{``populace''} by the model (OCV 39:448). This same label recurs in a page of the article \textit{``Esprit''} (spirit, wit), which the human indexer has again categorised as \textit{``style''}; the model adds \textit{``populace''} and \textit{``traduction''} (translation) --- the latter because Voltaire mentions the term \textit{``traduction''} in passing, merely to observe that only bad Latin translations of Hebrew writings use the word \textit{spiritus} in the sense of an ethereal being 
\parencite[p.~244]{voltaire_ocv_41}. The question of how to handle quotations is not straightforwardly resolved. On the one hand, quotations are tagged as such in the TEI-XML files underlying our experiments, and it would be technically feasible either to mark them as a feature or to exclude them from the text passed to the model. On the other hand, quotations do not always serve merely illustrative purposes: Voltaire frequently includes them for their content, in which case they are thematically relevant and ought to be taken into account. The distinction is usually clear to a human indexer but remains difficult to encode.

When the models were applied to corpora for which no index previously existed --- QE labels to the DP and the \textit{Lettres philosophiques}, EM labels to the \textit{Histoire de Charles~XII} --- performance declined appreciably, though in the absence of pre-existing human-generated labels any evaluation was necessarily subjective. The choice of source label set matters considerably: applying EM labels to the \textit{Histoire de Charles~XII} produced assignments that skewed heavily towards broad geographical headwords (\textit{``Russie''}, \textit{``Pologne''}, \textit{``Suède''}, \textit{``Estonie''}, \textit{``Finlande''}), which are of limited utility to a reader seeking to navigate the narrative and analysis of Charles~XII's campaigns. As has long been recognised in indexing theory, one of the indexer's primary tasks is the selection of headwords that collectively capture the distinctive thematic profile of the work in question \parencite[pp.~259--60]{duncan2022}; a label set designed for one work will almost inevitably prove ill-fitting for another, even when the two are generically related. The cross-corpus experiments described in Section~\ref{sec:final_ml} confirm this at the quantitative level: the 10--20 percentage point drop in F1 under out-of-distribution conditions reflects not only the change in text but the mismatch between label sets.

The DP article \textit{``Dieu''} offers a useful illustration of the cumulative difficulties the model faces when working outside its training distribution. Cast as a dialogue between Logomacos, a theologically sophisticated visitor from Constantinople, and Dondindac, a simple provincial believer from the south of France, the article interweaves fiction, philosophy, and polemic in ways that resist straightforward thematic assignment \parencite[pp.~21--28]{voltaire_ocv_36}. As Logomacos interrogates Dondindac, the model produces a sequence of labels --- \textit{``athéisme''} for a page about the practice of prayer; \textit{``ignorance''} for Dondindac's account of what he knows about God; \textit{``philosophie''} for a page in which Dondindac protests incomprehension before increasingly abstruse theological questions; \textit{``déisme''} where Dondindac does indeed profess deistic beliefs; \textit{``anthropomorphisme''} for a parable about a mole in dialogue with a mayfly; and \textit{``athéisme''} again for a page of 16 words that can hardly be said to have a discernible theme at all. While \textit{``déisme''} is plausible, the others are not, and the repeated assignment of \textit{``athéisme''} to pages with little or no connection to atheism is difficult to account for. In the article \textit{``Destin''}, for example, this label \textit{``athéisme''} is assigned to a page effectively limited to a cross-reference: \textit{``plutôt, si vous pouvez examiner paisiblement avec moi ce que c'est, passez à la lettre L.''} \parencite[p.~19]{voltaire_ocv_36}.

It is worth stepping back from the qualitative texture of individual results to reflect on a more fundamental question of method. Technological innovations have historically played a constitutive role in the genesis of indexing as a practice: it was only with the emergence of the printed book, with its multiple identically paginated copies, that producing an index became a worthwhile investment of labour and funds relative to the manuscript paradigm that preceded it \parencite[p.~466; 474]{abbamonte2023}. The index, in turn, gave rise to a new mode of engagement with text --- one in which readers could navigate selectively, following their own paths through a work according to a consultation model rather than reading in linear fashion from start to finish \parencite[pp.~117--21]{blair2010}. In the experiments described here, we retained the page as the unit of analysis, both to maintain compatibility with the print indexes and to allow the human-generated labels to serve as a control. Yet from a semantic perspective, the page is an unsatisfactory unit of analysis: again and again, examining the results, we encountered pages containing only the beginning or end of a longer passage whose subject cannot be reliably identified without the surrounding context. The codex, with its individually numbered pages, made indexing possible; but in a digital environment, there is no technical reason to be bound by this artefact. Indexing at the level of the paragraph --- or of the semantically coherent passage, as identified by the model itself --- would be both more natural and more accurate. This is perhaps the most significant methodological implication of our experiments, and one that points towards a genuinely different approach to digital indexing in future work.

We conclude this discussion with a final comment on the limitations of this study. Firstly, we have not carried out an uncertainty analysis of the numerical results presented; therefore, these should be regarded as indicative pending such an analysis. Secondly, and as we have previously pointed out, the main methodological variations considered in this project have been carried out simultaneously (e.g. change in model scale, encoder versus generative formulation, etc.), so each set of results must be considered valid within its own context, and comparisons should not be made that may be subject to confounding factors. Only in cases where we have expressly indicated a variation in a single condition (e.g. a change of corpus only, or a reduction in labels only) can firm conclusions be drawn regarding these specific changes. These limitations point to interesting avenues for future work, to which we add the extension of these methodologies to other corpora to confirm their generalisability.

\section{Conclusions}\label{sec:conclusions}

The experiments reported here demonstrate both the promise and the current limits of machine learning as a tool for automatic thematic indexing of large-scale literary and historical corpora. Fine-tuned generative LLMs --- and in particular the 4-bit quantised Mistral-Small-3.2-24B model --- achieved the best results in our experiments, with F1 scores of 0.64 and 0.43 on the QE and EM datasets respectively, rising to 0.67 and 0.47 when low-frequency labels are filtered out. These figures represent lower bounds on real performance: as the qualitative analysis makes clear, the automatic evaluation framework penalises semantically valid predictions that happen to differ from the human-generated ground truth, and manual inspection consistently reveals a proportion of ``false'' generated labels that are in fact plausible or even superior alternatives to the print index labels. However, it is worth noting that we have not quantified the difference between this lower bound and the potential average growth, and therefore we cannot draw any conclusions about the actual impact of this variation.

The qualitative analysis is, in many respects, the more revealing contribution of this study. Quantitative metrics capture what the model gets wrong relative to a fixed standard; the close reading of individual results illuminates \textit{why} it goes wrong and what that tells us about the nature of the task. The findings point consistently towards a cluster of difficulties that are unlikely to be resolved by scaling alone: the heavy skew of label distributions towards rare terms, the semantic richness and generic diversity of the source texts, and above all the rhetorical complexity of Voltaire's prose --- his irony, his misdirection, his habit of embedding serious argument in playful or fictional frames. These are not incidental features of the corpus but constitutive ones, and they suggest that the gap between model performance and professional indexing practice reflects something deeper than a lack of training data or model capacity. As \textcite{duncan2022} observes, ''every subject index worth its salt must, in an important sense, be a work of the imagination'' (p.~259). As such, the index is in many ways the product of interpretative judgements that current models can approximate but not yet replicate.

It is also worth reflecting on our choice of corpus. Voltaire was a natural starting point for this work, given the existence of the new digital editions of the OCV, complete with some professionally indexed volumes and the desirability of extending reliable thematic access to the unindexed works within and beyond it. He may not, however, be the easiest author with whom to begin. His tendency to weave disparate topics together, to write allusively, to deploy irony as a primary argumentative instrument, and to mean something rather different from what he says --- all qualities that were irresistible to his eighteenth-century readers --- make him a demanding subject for any system that relies on surface-level semantic signals. That the models perform as well as they do under these conditions is, on reflection, encouraging. All the more reason, then, to want to open up more paths into his vast body of writing --- but which is perhaps not the most forgiving place to begin developing a fledgling approach to automatic indexing.

Perhaps the most significant methodological lesson of these experiments concerns not the models themselves but the unit of analysis. As argued in the Discussion, the page --- inherited from the print tradition and retained here for reasons of comparability --- is a poor semantic unit for indexing purposes. A move towards paragraph- or passage-level indexing, exploiting the structural markup already present in the TEI-XML source files, would both improve model performance and produce indexes more naturally suited to digital navigation. Combined with the extension of trained models to unindexed works --- the original motivation for this project --- such a shift would represent a meaningful step towards the kind of genuinely structured access to large literary corpora that full-text search alone cannot provide. The problem of making Voltaire's 15 million words navigable to the non-specialist reader remains open; these experiments suggest that it is tractable, and indicate where the most productive directions for future work might lie.

\begin{acknowledgement}
The authors would like to thank Digital Scholarship @ Oxford (DiSc) for funding and supporting this project, with the particular involvement of Andrew Cusworth (DiSc and Bodleian Libraries, University of Oxford). Access to the digitised files of Voltaire's text with page numbers and the printed index labels was provided by the Voltaire Foundation (University of Oxford). We would also like to thank Haris Zia, who contributed to the project in its early stages.
\end{acknowledgement}

\begin{funding}
This project has been partially funded by Digital Scholarship @ Oxford (DiSc).
\end{funding}

\begin{conflictofinterest}
The authors state no conflict of interest.
\end{conflictofinterest}

\articlenote{%
\textbf{Data availability statement:} The software developed during this project, as well as the machine learning models we have trained, are publicly available as open-source in the following repository: \url{https://github.com/Digital-Scholarship-Oxford/digitally-indexing-voltaire}}

% \bibliographystyle{apalike}
% \bibliography{Bibliography}
\printbibliography
\end{document}